\documentclass[conference]{IEEEtran}

\usepackage{algorithm}
\usepackage{algpseudocode}
\usepackage{amsmath}
\usepackage{amssymb}
\usepackage{graphicx}
\usepackage{fancyhdr}
\usepackage{multirow}
\usepackage{threeparttable}

\begin{document}
	
\title{Parameter Sensitivity Analysis of Social Spider Algorithm}
	
\author{James J.Q. Yu,
	\textit{Student Member, IEEE} and
	Victor O.K. Li,
	\textit{Fellow, IEEE}\\
	Department of Electrical and Electronic Engineering\\
	The University of Hong Kong\\
	Email: \{jqyu, vli\}@eee.hku.hk\\
}

\maketitle
\thispagestyle{plain}
\fancypagestyle{plain}{
	\fancyhf{}      
	\fancyfoot[L]{978-1-4799-7492-4/15/\$31.00~\copyright2015~IEEE}
	\fancyfoot[C]{}
	\fancyfoot[R]{}
	\renewcommand{\headrulewidth}{0pt}
	\renewcommand{\footrulewidth}{0pt}
}

\pagestyle{fancy}{
	\fancyhf{}
	\fancyfoot[R]{}}
\renewcommand{\headrulewidth}{0pt}
\renewcommand{\footrulewidth}{0pt}
\pagestyle{empty}

\begin{abstract}
	Social Spider Algorithm (SSA) is a recently proposed general-purpose real-parameter metaheuristic designed to solve global numerical optimization problems. This work systematically benchmarks SSA on a suite of 11 functions with different control parameters. We conduct parameter sensitivity analysis of SSA using advanced non-parametric statistical tests to generate statistically significant conclusion on the best performing parameter settings. The conclusion can be adopted in future work to reduce the effort in parameter tuning. In addition, we perform a success rate test to reveal the impact of the control parameters on the convergence speed of the algorithm.
\end{abstract}

\begin{IEEEkeywords}
	Social spider algorithm, global optimization, parameter sensitivity analysis, evolutionary computation, meta-heuristic.
\end{IEEEkeywords}

\section{Introduction}
	Social Spider Algorithm (SSA) is a recently proposed meta-heuristic designed to solve real-parameter black-box optimization problems \cite{YuLi2015SocialSpiderAlgorithm}. SSA mimics the foraging behavior of the social spiders \cite{Uetz1992Foragingstrategiesspiders} for optimization. It has been applied in solving both benchmark optimization problems \cite{YuLi2015SocialSpiderAlgorithm} and real-world problems \cite{YuLi2014BaseStationSwitching}\cite{YuLiPowerControlledBase} and demonstrated satisfactory performance. However, a parameter sensitivity analysis on the control parameters of SSA is still lacking. Although the optimization performance of SSA in existing work outperforms the compared algorithms, it is significantly influenced by the four control parameters, i.e., population size, attenuation rate, change rate, and mask rate. Thus tuning these parameters properly is essential to demonstrate the efficacy of SSA as a problem solver.
	
	In general, the success of evolutionary algorithms (EAs) in solving optimization problems relies on the proper selection of control parameters \cite{AletiMoserMeedeniya2014ChoosingAppropriateForecasting}. Control parameter tuning is the process where different combinations of parameters are assessed to achieve a best performance in terms of solution quality and/or computational complexity. Parameter sensitivity analysis methods and applications have attracted much research attention in the past decades and many work has been done to fully exploit the searching power of EAs \cite{SmitEiben2009ComparingParameterTuning}. An early example of such applications was conducted by Grefenstette \cite{Grefenstette1986Optimizationcontrolparameters}, in which the control parameters of a genetic algorithm was tuned by another one so as to improve the performance on an image registration problem. Ever since then, parameter sensitivity tests have been performed on most of the existing EAs, including but not limited to PSO \cite{ShiEberhart1998ParameterSelectionin} and DE \cite{QinLi2013DifferentialEvolutionCEC}.
	
	In this work, we systematically study the impact of different parameter settings on a benchmark suite with a collection of optimization functions with different properties. We employ a set of 11 benchmark functions selected from a widely used benchmark function suite \cite{YaoLiu1997Fastevolutionstrategies} and from the recent CEC single-objective continuous optimization testbed \cite{LiangQuSuganthanHernandez-Diaz2013ProblemDefinitionsand}. All test functions are the base functions from these two sources. We evaluate SSA using a wide range of parameter settings on the 11 benchmark functions and present the best achieved results among all possible parameter combinations. In addition, we conduct a parameter sensitivity analysis with respect to all the combinations. The Friedman test and the Hochberg post-hoc procedure \cite{HollanderWolfe1999NonparametricStatisticalMethods}\cite{ShilaneMartikainenDudoitOvaska2008generalframeworkstatistical} are employed to generate statistically significant conclusion. The conclusion can be adopted in future related work to reduce the parameter tuning effort.
	
	The rest of this paper is organized as follows. Section II gives a brief review of SSA. Section III presents the experimental settings and simulation results. Section IV concludes this paper.
	
\section{Social Spider Algorithm}

	A real-parameter black-box optimization problem aims to find the optimal decision variables, or the global best solution, to minimize/maximize one or multiple objective functions. In the optimization process, no knowledge of the function characteristics is known to the problem solver. SSA is a population-based general-purpose stochastic optimization problem solver and has been shown to be effective and efficient in solving real-parameter black-box optimization problems \cite{YuLi2015SocialSpiderAlgorithm}. It has also been applied in scientific and engineering research \cite{YuLi2014BaseStationSwitching}\cite{YuLiPowerControlledBase}.
	
	In SSA, the search space of an optimization problem is considered as a hyper-dimensional spider web, on which a number of artificial spiders can move freely. Each position on the web corresponds to a feasible solution to the optimization problem. In addition, the web also serves as a transmission media for the vibrations generated by the movement of the spiders. Details of SSA will be introduced in the following subsections.
	
\subsection{Spider}
	
	The artificial spiders are the basic operating agents of SSA. Each spider possesses a position on the hyper-dimension spider web, and the fitness value of this position is assigned to the spider. Each spider holds a memory storing its status as well as optimization parameters, namely, its current position, current fitness value, following vibration at previous iteration, inactive degree, previous movements, and dimension mask. All these pieces of information are utilized to guide the spider to search for the global optimum, and will be introduced later.

\subsection{Vibration}

	In SSA, a spider will generate a vibration upon moving to a new position different from the one in the previous iteration. Vibration is a unique feature of SSA, which distinguishes SSA from other meta-heuristics. The vibrations serve as medium for the spiders to perform lossy communications with others to form a collective social knowledge.
	
	A vibration in SSA is characterized by two properties, namely, its source position and source intensity. When a spider moves to a new position $P$, it generates a vibration whose source position is $P$. The source intensity of this vibration is calculated using the following equation:
	\begin{equation}\label{eqn:intensity}
	I=\log(\frac{1}{f(P)-C}+1),
	\end{equation}
	where $I$ is the source intensity, $f(P)$ is the fitness value of $P$, and $C$ is a constant smaller than the infimum of the objective function\footnote{We consider minimization problem in this paper. For maximization problems, $C$ is a constant larger than the supremum of the objective function, and the denominator of the equation is $C-f(P)$.}.
	
	After a vibration is generated, it will propagate over the spider web. As a form of energy, vibration attenuates over distance. This physical phenomenon is considered in SSA. The vibration attenuation process is defined with the following equation:
	\begin{equation}\label{eqn:attenuation}
	I^{d}=I\times\exp(-\frac{d}{\overline{\sigma}\times r_a}),
	\end{equation}
	where $I^{d}$ is the attenuated intensity after propagating for distance $d$, $\overline{\sigma}$ is the average of the standard deviation of all spider positions along each dimension. In this equation, $r_a\in(0,+\infty)$ is introduced to control the attenuation rate of the vibration intensity, and will be tested as a user-controlled parameter in this paper.

\subsection{SSA Search Pattern}

	SSA manipulates a population of artificial spiders via a series of optimization step. Specifically, each iteration of SSA can be divided into the following steps.
	
\subsubsection{Fitness Evaluation}

	At the beginning of each iteration, the fitness values of the positions possessed of all spiders in the population is re-evaluated. These fitness values are later utilized in the vibration generation and propagation process.
	
\subsubsection{Vibration Generation}

	After all fitness values are calculated, each spider in the population will generate a new vibration at its current position. Its source intensity is calculated using (\ref{eqn:intensity}). This vibration is immediately propagated to all the spiders in the population via the spider web with attenuation. Thus, each spider will receive $|\textit{pop}|$ different vibrations from the web when all spiders have generated new vibrations in the current iteration, where $|\textit{pop}|$ is the population size. Upon the receipt of these $|\textit{pop}|$ vibrations, the spider will select the one with the largest attenuated vibration intensity, and compare it with the one it stored, called the ``following vibration'' in the previous iteration. The vibration with a larger intensity is stored as the updated following vibration in the current iteration. If the spider does not change its following vibration, its inactive degree is incremented by one. Otherwise this degree is reset to zero. This ends the vibration generation step.
	
\subsubsection{Mask Changing}

	When all spiders have determined their following vibration, the algorithm proceeds to manipulate their positions on the spider web. Here SSA utilizes a dimension mask to introduce randomness in the following movement process. Each spider holds a dimension mask which is a 0-1 binary vector of length $D$, where $D$ is the number of dimensions of the optimization problem. In each iteration, the spider has a probability of $1-p_c^{d_\textit{in}}$ to change its mask where $p_c\in(0,1)$ is a user-controlled parameter this paper aims to study. Here we use $d_\textit{in}$ to denote the inactive degree of the spider. If the mask is decided to be changed, each bit of the mask has a probability of $p_m$ to be assigned with a one, and $1-p_m$ to be a zero. $p_m$ is another user-controlled parameter in the parameter sensitivity analysis. In this process, each bit is changed independently, except for the case where all bits are set to zero. In such a case, one random bit of the mask is set to one in order to avoid premature convergence.
	
\subsubsection{Random Walk}
	
	With the following vibration and dimension mask determined, each spider then generates a following position based on these two pieces of information and all received vibrations. This following position is employed together with the spider's current position to generate a new position on the spider web. This new position is assigned to the spider and this ends the random walk step. A more detailed formulation can be found in \cite{YuLi2015SocialSpiderAlgorithm}.
	
\subsection{Control Parameters}

	In SSA, four user-controlled parameters shall be fine-tuned in order to achieve good performance:
	
\subsubsection{Population Size $|\textit{pop}|$}

	The size of the spider population determines the individual diversity and influences the convergence speed. The selection of this parameter greatly influences the computational complexity as the standard deviation of all solutions shall be calculated when generating attenuated vibration intensities using (\ref{eqn:attenuation}). In addition, while a larger population may increase the diversity of the population which in turn increases the exploration ability, a smaller population may result in better convergence speed.
	
	\begin{table*}[t]
		\caption{Benchmark Functions}
		\label{tbl:benchmark}
		\begin{center}
			\begin{threeparttable}
				\begin{tabular}{lccl}
					\hline
					Function & Search Range & Dimension ($D$) & Name \\
					\hline
					$\begin{aligned}f_1(\textbf{x})=\sum_{i=1}^{D}x_i^2\end{aligned}$ & $[-100,100]^D-\textbf{o}$ & 30 & Generalized Sphere Function \\
					$\begin{aligned}f_2(\textbf{x})=x_1^2\sum_{i=2}^{D}10^6x_i^2\end{aligned}$ & $[-100,100]^D-\textbf{o}$ & 30 & Generalized Cigar Function \\
					$\begin{aligned}f_3(\textbf{x})=\sum_{i=1}^{D}(\sum_{j=1}^{i}x_i)^2\end{aligned}$ & $[-100,100]^D-\textbf{o}$ & 30 & Schwefel's Function 1.2 \\
					$\begin{aligned}f_4(\textbf{x})=\max_i{|x_i|}\end{aligned}$ & $[-100,100]^D-\textbf{o}$ & 30 & Schwefel's Function 2.21 \\
					\hline
					$\begin{aligned}f_5(\textbf{x})=\sum_{i=1}^{D-1}(100(x_{i+1}-x_i^2)^2+(x_i-1)^2)\end{aligned}$ & $[-100,100]^D-\textbf{o}$ & 30 & Generalized Rosenbrock's Function \\
					$\begin{aligned}
					f_6(\textbf{x})=&418.9829D-\sum_{i=1}^{D}g(x_i+420.9687)\\
					&g(z_i)=
					\begin{cases}
					z_i\sin(|z_i|^{1/2}) & \mathrm{if\:}|z_i|<500 \\
					y_i\sin(\sqrt{|y_i|})-\frac{(z_i-500)^2}{10^4D} & \mathrm{if\:}z_i>500 \\
					y_i\sin(\sqrt{|y_i|})-\frac{(z_i+500)^2}{10^4D} & \mathrm{if\:}z_i<-500
					\end{cases} \\
					&y_i=
					\begin{cases}
					500-\text{mod}(z_i,500) & \mathrm{if\:}z_i>500 \\
					\text{mod}(-z_i,500)-500 & \mathrm{if\:}z_i<-500
					\end{cases} \\
					\end{aligned}$ & $[-1000,1000]^D-10\textbf{o}$ & 30 & Modified Schwefel's Function \\
					$\begin{aligned}f_7(\textbf{x})=\sum_{i=1}^{D}(x_i^2-10\cos(2\pi x_i)+10)\end{aligned}$ & $[-5.12,5.12]^D-0.0512\textbf{o}$ & 30 & Generalized Rastrigin's Function \\
					$\begin{aligned}
					f_8(\textbf{x})=&-20\exp(-0.2\sqrt{\frac{1}{n}\sum_{i=1}^nx_i^2})-\exp[\frac{1}{n}\sum_{i=1}^n\cos(2\pi x_i)]\\&+20+e
					\end{aligned}$ & $[-32,32]^D-0.32\textbf{o}$ & 30 & Ackley's Function \\
					$\begin{aligned}f_9(\textbf{x})=\frac{1}{4000}\sum_{i=1}^nx_i^2-\prod_{i=1}^n\cos(\frac{x_i}{\sqrt{i}})+1\end{aligned}$ & $[-600,600]^D-6\textbf{o}$ & 30 & Generalized Griewank's Function \\
					$\begin{aligned}f_{10}(\textbf{x})=&g(x_1,x_2)+g(x_2,x_3)+\cdots+g(x_{n-1},x_n)+g(x_n,x_1)\\&g(x,y)=0.5+\frac{(\sin^2(\sqrt{x^2+y^2})-0.5)}{(1+0.001(x^2+y^2))^2}\end{aligned}$ & $[-100,100]^D-\textbf{o}$ & 30 & Scaffer's Function F6 \\
					$\begin{aligned}f_{11}(\textbf{x})=\frac{10}{D^2}\prod_{i=1}^D(1+i\sum_{j=1}^32\frac{|2^jx_i-\text{round}(2^jx_i)|}{2^j}^{\frac{10}{D^{1.2}}}-\frac{10}{D^2}\end{aligned}$ & $[-5.12,5.12]^D-0.0512\textbf{o}$ & 30 & Katsuura's Function \\
					\hline
				\end{tabular}
			\end{threeparttable}
		\end{center}
	\end{table*}
	
\subsubsection{Attenuation Rate $r_a$}

	This parameter controls the attenuation strength imposed on the vibration intensities. A larger $r_a$ will make the exponential part of (\ref{eqn:attenuation}) approach one, resulting in a weaker attenuation on the vibration intensity. This means that the communication is less lossy and best positions in the population are more likely to be broadcast to all spiders. On the contrary, a smaller $r_a$ results in a stronger attenuation, rendering the information propagation more lossy. This may potentially obstruct the algorithm from fast convergence but prevent it from premature convergence.
	
\subsubsection{Mask Changing Probabilities $p_c$ and $p_m$}

	These two parameters cooperate to determine the dimension mask of each spider in the population. A large $p_c$ will result in frequent mask changing, which can be beneficial to solve optimization problems with a great number of local optima. A large $p_m$ will result in more 1's in the dimension mask, which means that more noise is introduced in the following position when performing random walk \cite{YuLi2015SocialSpiderAlgorithm}.
	
\section{Experimental Settings and Simulation Results}

	We evaluate the optimization performance of SSA with all possible 900 combinations of six $|\textit{pop}|$ values, six $r_a$ values, five $p_c$ values, and five $p_m$ values on 11 numerical test functions selected from the base functions of a widely-adopted benchmark suite \cite{YaoLiu1997Fastevolutionstrategies} and a recent CEC single-objective continuous optimization testbed \cite{LiangQuSuganthanHernandez-Diaz2013ProblemDefinitionsand}. These functions are employed to construct the competition functions in recent CEC single-objective optimization competitions \cite{LiangQuSuganthanHernandez-Diaz2013ProblemDefinitionsand}\cite{SuganthanHansenLiangDebChenAugerTiwari2005ProblemDefinitionsand}\cite{LiangQuSuganthan2014ProblemDefinitionsand}, and are considered representative for covering most possible features of real-world application problems. In addition, we report the simulation results of the best performing combination of control parameters with additional performance analysis. To provide a statistically significant and reliable guideline for parameter selection, we perform the parameter sensitivity analysis using a rigid non-parametric statistical test framework to compare all 900 test cases.
	
\subsection{Benchmark Functions}

	In this work, we adopt 11 base functions from \cite{YaoLiu1997Fastevolutionstrategies} and from the CEC 2013 single-objective continuous optimization testbeds. These base functions are grouped into two classes: uni-modal functions $f_1$--$f_4$ and multi-modal functions $f_5$--$f_{11}$. The benchmark functions are listed in Table \ref{tbl:benchmark}.
	
	As one of the objective is to test the convergence ability of the algorithm, we use a novel method to shift the position of the global optima yet maintain the accuracy of computation to smaller than $10^{-16}$ \cite{SuganthanHansenLiangDebChenAugerTiwari2005ProblemDefinitionsand}. Previously, most benchmark function suites move the global optima from $\textbf{0}$ to a random position in order to avoid unfair competition by algorithms that only exploits the center of the search space. However, due to the storage method in modern computer, the accuracy of a floating point number is limited to its size, which makes it difficult to store large numbers with small tails, e.g. $\alpha=50+10^{-50}$. In order to preserve a high accuracy and avoid global optima at $\textbf{0}$, the benchmark functions in this paper adopt a search space shift vector $\textbf{o}$ on the search spaces. For example, $f_1$ originally is designed to search for global optimum in $[-100,100]^D$. In our benchmark function suite, the search space for $f_1$ is modified to $[-100,100]^D-\textbf{o}=[-100-o_1,100-o_1]\times[-100-o_2,100-o_2]\times\cdots\times[-100-o_D,100-o_D]$. By this means, the global optimum, although still at $\textbf{0}$, is not the center of the search space anymore. The value of $\textbf{o}$ is set according to \cite{SuganthanHansenLiangDebChenAugerTiwari2005ProblemDefinitionsand}.
	
\subsection{Experimental Settings}

	It is a very time consuming task to choose a proper parameter configuration for SSA to generate outstanding performance. As many real-world application problems are computationally expensive, it may take much time to finish one run of the algorithm, which makes the trial-and-error method for parameter tuning impractical. One solution to this problem is to employ the ``fixed scheme'' for parameter configuration, which uses a fixed parameter setting throughout the search based on extensive empirical studies on a wide range of test functions with diverse properties. Such empirical studies can deduce guidelines for choosing parameter configurations leading to consistently satisfactory performance over most optimization problems.
	
	In this paper, we employ advanced non-parametric statistical tests to conduct parameter sensitivity analysis and to determine some parameter combinations that can lead to statistically significant better performance than others. Our statistical analysis is a two-step framework. In the first step, the Friedman test \cite{HollanderWolfe1999NonparametricStatisticalMethods}\cite{III2005Contributionstotwo} is conducted to compare 900 parameter settings over the 11 test functions to test the null hypothesis that no test case performs significantly different from others. If this null hypothesis is rejected, a Hochberg post-hoc procedure \cite{ShilaneMartikainenDudoitOvaska2008generalframeworkstatistical}\cite{DerracGarciaMolinaHerrer2011practicaltutorialuse} is adopted to determine which settings perform significantly better than others.
	
	We test SSA with 900 parameter combination settings, namely $[|\textit{pop}|, r_a, p_c, p_m]\in[10,20,30,40,50,70]\times[0.2,$ $0.5,1.0,2.0,4.0,8.0]\times[0.1,0.3,0.5,0.7,0.9]\times[0.1,0.3,0.5,$ $0.7,0.9]$ on all 11 benchmark functions on 30 dimensions, respectively. For each benchmark function, SSA with each of 900 parameter settings are executed 51 times. The stopping criteria for all runs are set to 300 000 function evaluations, according to the suggestion by \cite{LiangQuSuganthanHernandez-Diaz2013ProblemDefinitionsand}.

\subsection{Parameter Sensitivity Analysis}

	\begin{table}[t]
		\caption{Parameter Sensitivity Analysis Result}
		\label{tbl:sensitivity}
		\begin{center}
			\begin{threeparttable}
				\begin{tabular}{c|c|cccccc}
				\hline
				 & & \multicolumn{6}{c}{$r_a$} \\ \hline
				$p_m$ & $p_c$ & 0.2 & 0.5 & 1.0 & 2.0 & 4.0 & 8.0 \\
				\hline
				\multirow{5}*{0.1} & 0.1 & 20/30 & 20/30 & 20/40 & 20/30 & - & - \\
				 & 0.3 & 20/30 & 20/30 & 20/30/40 & 20/30 & - & - \\
				 & 0.5 & 10/20/30 & 20/30 & 20/30/40 & 20/30/40 & - & - \\
				 & 0.7 & 10/20/30 & 10/20/30 & 20/30/40 & 20/30/40 & - & - \\
				 & 0.9 & 10/20/30 & 10/20/30 & 10/20/30 & 20/30 & - & - \\ \hline
				 \multirow{5}*{0.3} & 0.1 & - & - & - & - & - & - \\
				 & 0.3 & - & - & - & - & - & - \\
				 & 0.5 & 20/30 & 20/30 & 20 & 20/30/40/50 & - & - \\
				 & 0.7 & - & - & - & 40 & - & - \\
				 & 0.9 & - & - & - & - & - & - \\ \hline
				 \multirow{5}*{0.5} & 0.1 & - & - & - & - & - & - \\
				 & 0.3 & - & - & - & - & - & - \\
				 & 0.5 & - & - & - & - & 30 & - \\
				 & 0.7 & - & - & - & - & - & - \\
				 & 0.9 & - & - & - & - & - & - \\ \hline
				 \multirow{5}*{0.7} & 0.1 & - & - & - & - & - & - \\
				 & 0.3 & - & - & - & - & - & - \\
				 & 0.5 & - & - & - & - & - & - \\
				 & 0.7 & - & - & - & - & - & - \\
				 & 0.9 & - & - & - & - & - & - \\ \hline
				 \multirow{5}*{0.9} & 0.1 & - & - & 70 & - & - & - \\
				 & 0.3 & - & - & - & - & - & - \\
				 & 0.5 & - & - & - & - & - & - \\
				 & 0.7 & - & 40/50/70 & - & - & - & - \\
				 & 0.9 & - & 50/70 & 70 & - & - & - \\
				\hline
				\end{tabular}
			\end{threeparttable}
		\end{center}
	\end{table}
	
	Table \ref{tbl:sensitivity} reports the results of the parameter sensitivity analysis. In the table, different $r_a$ values are presented horizontally while $p_c\times p_m$ combinations are presented vertically. If a specific $|pop|\times r_a \times p_c\times p_m$ combination performs significantly better than others, $|pop|$ is labeled in the corresponding $r_a \times p_c\times p_m$ cell, i.e., all parameter combination labeled in the table performs significantly better than those not labeled.
	
	From this table, we have the following observations:
	\begin{itemize}
		\item Medium $|pop|$ values (e.g., 20, 30, and 40), small to medium $r_a$ values (e.g., 0.5, 1, and 2), medium $p_c$ values (e.g., 0.5 and 0.7), and small $p_m$ values (e.g., 0.1 and 0.3) are preferred by the statistical analysis.
		\item A roughly most preferred parameter combination for solving the benchmark functions in this paper is $[|pop|, r_a, p_c, p_m]=[30/1.0/0.7/0.1]$. This result accords with our preliminary observation in \cite{YuLi2015SocialSpiderAlgorithm}.
		\item There are some noisy data points at $p_m=0.9$. After a further investigation of the raw simulation data, we found that the parameter combinations with $p_m=0.9$ sacrifices the performance on uni-modal functions for some tiny improvements in $f_{11}$, which is multi-modal, non-separable, and the number of local optima is huge. As the improvement is barely enough to outperform others, it possesses advantage in the statistical analysis. However, they are not preferred when considering the overall performance on all benchmark functions.
	\end{itemize}
	
	The simulation results for the recommended parameter combination by the statistical test are listed in Table \ref{tbl:recommendedresult}.
	
	\begin{figure*}[t]
		\includegraphics[width=\linewidth]{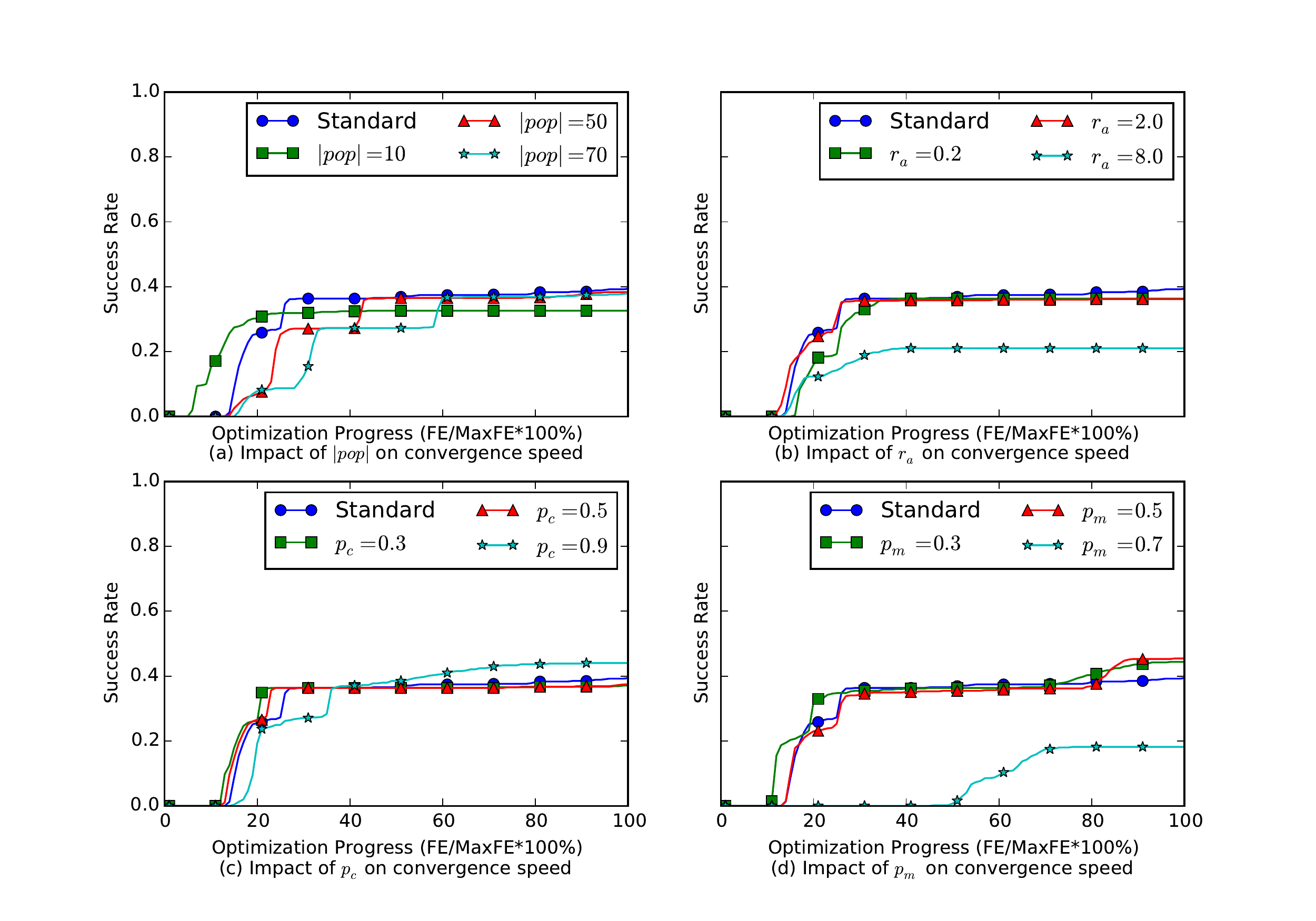}
		\caption{Success Rate Plots of Different Parameters.}
		\label{fig:conv}
	\end{figure*}
	
	\begin{table*}[t]
		\caption{Parameter Sensitivity Analysis Result}
		\label{tbl:recommendedresult}
		\begin{center}
			\begin{threeparttable}
				\begin{tabular}{c|ccccc}
					\hline
					Function & Mean & Std. Div. & Best & Worst & Median \\
					\hline
					$f_1$ & 1.1321E-73 & 1.3464E-73 & 2.4108E-75 & 7.8390E-73 & 7.7373E-74 \\
					$f_2$ & 9.2887E-70 & 5.2257E-69 & 3.3337E-78 & 3.7499E-68 & 4.3240E-74 \\
					$f_3$ & 1.3651E+00 & 8.9476E-01 & 1.5132E-01 & 3.8509E+00 & 1.1759E+00 \\
					$f_4$ & 3.9103E-03 & 4.4044E-03 & 2.7803E-04 & 2.5231E-02 & 2.1600E-03 \\
					$f_5$ & 6.2745E+01 & 1.0091E+02 & 4.0258E-04 & 5.1402E+02 & 1.8734E+01 \\
					$f_6$ & 1.9676E+01 & 1.1145E+02 & 1.6694E-01 & 8.0221E+02 & 1.4719E+00 \\
					$f_7$ & 6.1569E-07 & 4.1956E-06 & 0.0000E+00 & 3.0274E-05 & 0.0000E+00 \\
					$f_8$ & 7.4798E-15 & 4.9258E-16 & 3.9968E-15 & 7.5495E-15 & 7.5495E-15 \\
					$f_9$ & 6.2035E-13 & 1.8712E-12 & 0.0000E+00 & 1.2479E-11 & 3.9191E-14 \\
					$f_{10}$ & 6.4117E+00 & 3.9909E-01 & 5.2466E+00 & 7.1609E+00 & 6.4479E+00 \\
					$f_{11}$ & 2.1116E-01 & 1.3611E-01 & 2.5308E-03 & 4.5302E-01 & 2.2008E-01 \\
					\hline
				\end{tabular}
			\end{threeparttable}
		\end{center}
	\end{table*}
	
	Besides the numerical analysis on the simulation results, convergence speed is also a critical performance metric for meta-heuristics. In order to investigate the impact of parameter settings on the convergence speed, we adopt the success rate analysis \cite{QinLi2013DifferentialEvolutionCEC} to see the influence of the four parameters. In this success rate analysis, a simulation run is considered successful when the best fitness value is smaller than $10^{-8}$. For each run, the best fitness value is recorded every 3000 FEs and there are in total 100 ``success bits'' per run, recorded whether the run is successful up to the specific FEs.  Thus 576 ($11\times51$) runs of success bits for each parameter setting can be accumulated and presented as a plot of success rate over all test functions.
	
	In this test we employ the previously recommended parameter setting $[30/1.0/0.7/0.1]$ as the standard setting, denoted as ``Standard''. Different values of the four parameters are employed to compare with the standard setting, and only one parameter is consider in each comparison. The success rate plots are shown in Figure \ref{fig:conv}. In this figure, plot (a) studies the impact of $|pop|$ values on the convergence speed, plot (b) studies the impact of $r_a$, plot (c) studies the impact of $p_c$, and plot (d) studies the impact of $p_m$. From this figure, we have the following observations:
	
	\begin{itemize}
		\item The impact of $|pop|$ on the convergence speed is roughly the most significant one. This observations accords with our previous analysis in Section II-D.
		\item Reasonable $r_a$, $p_m$, and $p_c$ values yield similar convergence speed. For instance, the convergence plots for $r_a\in[1.0,2.0]$ are almost identical for the whole process, and the plot for $r_a=0.2$ joins them at approximately 35\% of the whole optimization process. However, $r_a=8.0$, which is a very rare setting for this parameter, cannot reach the success rates of others for the whole process.
	\end{itemize}
	
\section{Conclusion}

In this work we systematically evaluate the performance of SSA on a benchmark suite of 11 different functions. We carry out a parameter sensitivity analysis on 900 possible parameter settings on the four control parameters ($[|\textit{pop}|, r_a, p_c, p_m]\in[10,20,30,40,50,70]\times[0.2,$ $0.5,1.0,2.0,4.0,8.0]\times[0.1,0.3,0.5,0.7,0.9]\times[0.1,0.3,0.5,$ $0.7,0.9]$) using a rigid statistical test framework. The results indicate that medium $|pop|$ values (e.g., 20, 30, and 40), small to medium $r_a$ values (e.g., 0.5, 1, and 2), medium $p_c$ values (e.g., 0.5 and 0.7), and small $p_m$ values (e.g., 0.1 and 0.3) can lead to statistically significantly better performance than other settings. This is a reliable rule of thumb for parameter tuning of SSA in the future work. In addition, we using the success rate test to analyze the impact of these four parameters on the convergence speed of SSA. The simulation results indicate that while $|pop|$ value has a significant influence, rational $r_a$, $p_m$, and $p_c$ values generally lead to similar convergence speed.
	
\bibliographystyle{IEEEtran}
\bibliography{IEEEabrv,../../../bib/publications}

\end{document}